\documentclass[11pt,a4paper]{article}
\usepackage{authblk}

\usepackage{naaclhlt2019}
\usepackage{times}
\usepackage{latexsym}
\usepackage{amsmath} 
\usepackage{amssymb}
\usepackage{graphicx}
\usepackage{subcaption}
\usepackage{booktabs}
\usepackage{url}

\aclfinalcopy 
\def\aclpaperid{2327} 

\title{Mining Discourse Markers\\for Unsupervised Sentence Representation Learning
}

\author[3,1]{Damien Sileo}
\author[2]{Tim Van De Cruys}
\author[1]{Camille Pradel}
\author[3]{Philippe Muller}

\affil[1]{Synapse D\'eveloppement, Toulouse, France}
\affil[2]{IRIT, CNRS, France} \vspace{-2ex}
\affil[3]{IRIT, University of Toulouse, France}
\affil[ ]{\url{damien.sileo@synapse-fr.com}}

\vspace{-2ex}

\date{}

\begin{document}
\maketitle
\vspace*{-2em}
\begin{abstract}
Current state of the art systems in NLP heavily rely on manually annotated datasets, which are expensive to construct. Very little work adequately exploits unannotated data -- such as discourse markers between sentences -- mainly because of data sparseness and ineffective extraction methods. In the present work, we propose a method to automatically discover sentence pairs with relevant discourse markers, and apply it to massive amounts of data. Our resulting dataset contains 174 discourse markers with at least 10{\sc k} examples each, even for rare markers such as \textit{coincidentally} or \textit{amazingly}. We use the resulting data as supervision for learning transferable sentence embeddings. 
In addition, we show that even though sentence representation learning through prediction of discourse markers yields state of the art results across different transfer tasks, it is not clear that our models made use of the semantic relation between sentences, thus leaving room for further improvements.
Our datasets are publicly available \footnote{\url{https://github.com/synapse-developpement/Discovery}}

\end{abstract}
\section{Introduction}

An important challenge within the domain of natural language processing is the construction of adequate semantic representations
for textual units -- from words over sentences to whole documents. 
Recently, numerous approaches have been proposed for the construction of vector-based representations for larger textual units, especially sentences. 
One of the most popular frameworks aims to induce sentence embeddings as an intermediate representation for predicting relations between sentence pairs. For instance, similarity judgements (paraphrases) or inference relations have been used as prediction tasks, and the resulting embeddings perform well in practice, even when the representations are transfered to other semantic tasks \citep{Conneau2017}. However, the kind of annotated data that is needed for such supervised approaches is costly to obtain, prone to bias, and arguably fairly limited with regard to the kind of semantic information captured, as they single out a narrow aspect of the entire semantic content. 

Unsupervised approaches have also been proposed, based on sentence distributions in large corpora in relation to their discourse context.
For instance, \citet{Kiros2015} construct sentence representations by trying to reconstruct neighbouring sentences, which allows them to take into account different contextual aspects of sentence meaning. In the same vein, \citet{Logeswaran2016} propose to predict if two sentences are consecutive, even though such local coherence can be straightforwardly predicted with relatively shallow features \citep{Barzilay2008}.
A more elaborate setting is the prediction of the semantic or rhetorical relation between two sentences, as is the goal of discourse parsing. A number of annotated corpora exist, such as RST-DT \cite{DBLP:conf/sigdial/CarlsonMO01} and PDTB \cite{pdtb2.0}, 
but in general the available data is fairly limited, and the task of discourse relation prediction is rather difficult. 
The problem, however, is much easier when there is a marker that makes the semantic link explicit \citep{pitler2008}, and this observation has often been used in a semi-supervised setting to predict discourse relations in general \cite{RutherfordX15}. Building on this observation, one approach to learn sentence representations is to predict such markers or clusters of markers explicitly \citep{Jernite2017,Malmi2018,Nie2017}. 
Consider the following sentence pair:
 \begin{quote}
\textit{I live in Paris. But I'm often abroad.}
\end{quote}

\begin{table*}[htb]
\begin{tabular}{l|l}
s1 & Paul Prudhomme's Louisiana Kitchen created a sensation when it was published in 1984.\\
c  & happily,                                                                                                                          \\
s2' & This family collective cookbook is just as good                                                                                  
\end{tabular}
\caption{Sample from our \textit{Discovery} dataset}
\label{tab:sample}
\end{table*}

The discourse marker $\textit{but}$ highlights an opposition between the first sentence (the speaker lives in Paris) and the second sentence (the speaker is often abroad).
The marker can thus be straightforwardly used as a label between sentence pairs. In this case, the task is to predict $c=\textit{but}$ (among other markers) for the pair $(\textit{I live in Paris}, \textit{I'm often abroad})$.
Note that discourse markers can be considered as noisy labels for various semantic tasks, such as entailment ($c=\textit{therefore}$), subjectivity analysis ($c=\textit{personally}$) or sentiment analysis ($c=\textit{sadly}$). 
More generally, discourse markers indicate how a sentence contributes to the meaning of a text, and they provide an appealing supervision signal for sentence representation learning based on language use. 

 A wide variety of discourse usages would be desirable in order to learn general sentence representations. Extensive research in linguistics has resulted in elaborate discourse marker inventories for many languages.\footnote{See for instance a sample of language on the Textlink project website: \url{http://www.textlink.ii.metu.edu.tr/dsd-view}}
 These inventories were created by manual corpus exploration or annotation of small-scale corpora: the largest annotated corpus, the English PDTB 
 consists of a few tens of thousand examples, and provides a list of about 100 discourse markers, organized in a number of categories. 

\begin{table*}[htb]
    \begin{center}
    \begin{tabular}{lp{0.5\linewidth}cc}
    \toprule
    author  & discourse markers / classes                                                                     & classes & markers    \\ \midrule
    \citet{Jernite2017} & \sc{\small addition, contrast, time, result, specific, compare, strength, return, recognize}                       & 9  & 40 \\
    \citet{Nie2017}     & {\it and, but, because, if, when, before, though, so, as, while, after, still, also, then, although} & 15 & 15 \\
    current work      &  {\it later, often, understandably, gradually, or, ironically, namely, \ldots} & 174 & 174 \\ \bottomrule
    \end{tabular}                  
    \end{center}
       \caption{Discourse markers or classes used by previous work on unsupervised representation learning}
    \label{tab:previousconnectives}
\end{table*}

Previous work on sentence representation learning with discourse markers makes use of even more restricted sets of discourse markers, as shown in table \ref{tab:previousconnectives}. 
\citet{Jernite2017} use 9 categories as labels, accounting for 40 discourse markers in total. It should be noted that the aggregate labels do not allow for any fine-grained distinctions; for instance, the {\sc time} label includes both \textit{now} and \textit{next}, which is likely to impair the supervision. Moreover, discourse markers may be ambiguous; for example \textit{now} can be used to express contrast. On the other hand, \citet{Nie2017} make use of 15 discourse markers, 5 of which are accounting for more than $80 \% $ of their training data. In order to ensure the quality of their examples, they only select pairs matching a dependency pattern manually specified for each marker.
As such, both of these studies use a restricted or impoverished set of discourse markers; they also both use the BookCorpus dataset, whose size
\citep[$4.7M$ sentences that contain a discourse marker, according to][]{Nie2017} is prohibitively small for the prediction of rare discourse markers.

In this work we use web-scale data in order to explore the prediction of a wide range of discourse markers, with more balanced frequency distributions, along with application to sentence representation learning. 
We use English data for the experiments, but the same method could be applied to any language that bears a typological resemblance with regard to discourse usage, and has sufficient amounts of textual data available (e.g. German or French). 
Inspired by recent work  \citep{Dasgupta2018,Poliak2018, Levy2018, Glockner2018} on the unexpected properties of recent manually labelled datasets (e.g. SNLI), we will also analyze our dataset to check whether labels are easy to guess, and whether the proposed model architectures make use of high-level reasoning for their predictions. Our contributions are as follows:
\begin{itemize}
\item[--]  we propose a simple and efficient method to discover new discourse markers, and present a curated list of 174 markers for English;
\item[--]  we provide evidence that many connectives can be predicted with only simple lexical features;
\item[--]  we investigate whether relation prediction actually makes use of the relation between sentences;
\item[--]  we carry out extensive experiments based on the Infersent/SentEval framework.
\end{itemize}

\section{Discovering discourse markers}
\subsection{Rationale}
Our goal is thus to capture semantic aspects of sentences by means of distributional observations. For our training signal, we aim at something more evolved than just plain contextual co-occurrence, but simpler than a full-fledged encoder-decoder {\it \`{a} la} Skip-Thought. In that respect, discourse relations are an interesting compromise, if we can reliably extract them in large quantities. This objective is shared with semi-supervised approaches to discourse relation prediction, where automatically extracted explicit instances feed a model targetting implicit instances \cite{marcu2002unsupervised,DBLP:journals/nle/SporlederL08,DBLP:conf/acl/PitlerN09,RutherfordX15}. In this perspective, it is important to collect unambiguous instances of potential discourse markers. To do so, previous work used heuristics based on specific constructs, especially syntactic patterns for intra-sentential relations, based on a fixed list of manually collected discourse markers. Since we focus on sentence representations, we limit ourselves to discourse arguments that are well-formed sentences, thus also avoiding clause segmentation issues.  

Following a heuristic from \citet{RutherfordX15}, also considered by \citet{Malmi2018} and \citet{Jernite2017}, we collect pairs of sentences $(s_1, s_2)$ where $s_2$ starts with marker $c$. We only consider the case where $c$ is a single word, as detecting longer adverbial constructions is more difficult. We remove $c$  from the beginning of $s_2$ and call the resulting sentence $s_2'$. 
\citet{Malmi2018}  make use of a list of the 80 most frequent discourse markers in the PDTB in order to extract suitable sentence pairs.
We stay faithful to \citet{RutherfordX15}'s heuristic, as opposed to 
\citet{Malmi2018,Jernite2017}: if $s_2$ starts with $c$ followed by a comma, and $c$ is an adverbial or a conjunction, then it is a suitable candidate. By limiting ourselves to sentences that contain a comma, we are likely to ensure that $s_2'$ is meaningful and grammatical. As opposed to all the cited work mentioned above, we do not restrict the pattern to a known list of markers, but try to collect new reliable cues. 

This pattern is decisively restrictive, since discourse markers often appear at the clausal level (e.g. \textit{I did it but now I regret it}).
But clauses are not meant to be self contained, and it is not obvious that they should be included in a dataset for \textit{sentence} representation learning. At the same time, one could easily think of cases where $c$ is not a discourse marker, e.g. $(s_1, s_2)$= (``It's cold.'', ``Very, very cold.''). However, these uses might be easily predicted with shallow language models.
In the next section, we use the proposed method for the discovery of discourse markers, and we investigate whether the resulting dataset leads to improved model performance.

\subsection{Methodology}

We use sentences from the Depcc corpus \citep{Panchenko2017}, which consists of English texts harvested from commoncrawl web data. We sample 8.5 billion consecutive sentence pairs from the corpus.
We keep 53\% of sentence pairs that contain between 3 and 32 words, have a high probability of being English ($>75\%$) using FastText langid from \citet{GRAVE18.627}, have balanced parentheses and quotes, and are mostly lowercase. 
We use NLTK \citep{Bird2009} as sentence tokenizer and NLTK PerceptronTagger as part of speech tagger for adverb recognition.
In addition to our automatically discovered candidate set, we also include all (not necessarily adverbial) PDTB discourse markers that are not induced by our method. 
Taking this into account, 3.77\% of sentence pairs contained a discourse marker candidate, which is about 170{\sc m} sentence pairs. 
An example from the dataset is shown in table \ref{tab:sample}.
We only keep pairs in which the discourse marker occurs at least 10{\sc k} times. We also subsample pairs so that the maximum occurrence count of a discourse marker is 200{\sc k}.  The resulting dataset contains 19{\sc m} pairs. 

\begin{figure}[]
  \centering
\includegraphics[width =0.45\textwidth]{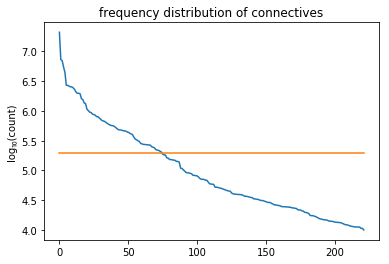}
  \caption{Frequency distribution of candidate discourse markers; 
  the horizontal line indicates the subsampling threshold.}
\label{fig:freq} 
\end{figure}

We discovered 243 discourse marker candidates. Figure \ref{fig:freq} shows their frequency distributions. As expected, the most frequent markers dominate the training data, but when a wide range of markers is included, the rare ones still contribute up to millions of training instances. 
Out of the 42 single word PDTB markers that precede a comma, 31 were found by our rule. Some markers are missing because of NLTK errors, which mainly result from morphological issues.\footnote{For instance, {\it lovely} is tagged as an adverb because of its suffix, while {\it besides} was never tagged as an adverb}

\subsection{Controlling for shallow features}
As previously noted, some candidates discovered by our rule may not be actual discourse markers. In order to discard them, we put forward the hypothesis that actual discourse markers cannot be predicted with shallow lexical features.
Inspired by \citet{Gururangan2018}, we use a Fasttext classifier \citep{joulin2016tricks} in order to predict $c$ from $s_2'$. The Fasttext classifier predicts labels from an average of word embeddings fed to a linear classifier. We split the dataset in 5 folds, and we predict markers for each fold, while training on the remaining folds. We use a single epoch, randomly initialized vectors of size 100 (that can be unigrams, bigrams or trigrams) and a learning rate of 0.5.

In addition, we predict $c$ from the concatenation of $s_1$ and $s'_2$ (using separate word representations for each case). One might assume that the prediction of $c$ in this case relies on the interaction between $s_1$ and $s_2$; however, the features of $s_1$ and $s_2$ within Fasttext's setup only interact additively, which means that the classification most likely relies on individual cues in the separate sentences, rather than on their combination.  In order to test this hypothesis, we introduce a \textit{random shuffle} operation: for each example ($s_1$, $s'_2$, $c$), $s'_2$ is replaced by a random sentence from a pair that is equally linked by $c$ (we perform this operation separately in train and test sets).

Table \ref{table:resshallowpred} indicates that shallow lexical features indeed yield relatively high prediction rates. Moreover, the shuffle operation indeed increases accuracy, which corroborates the hypothesis that classification with shallow features relies on individual cues from separate sentences, rather than their combination.

\begin{table}[htb]
\begin{center}
\begin{tabular}{lll}
\toprule
features         & accuracy ($\%$) \\ \midrule
majority rule    & 1.2             \\
s2               & 18.6           \\
s1-s2'            & 21.9         \\    
s1-s2' (shuffled) & 24.8              \\
\bottomrule
\end{tabular}
\end{center}
\caption{Accuracy when predicting candidate discourse markers using shallow lexical features}
\label{table:resshallowpred}
\end{table}

Tables \ref{table:lesspredictable} and \ref{table:mostpredictable} show the least and most predictable discourse markers, and the corresponding recognition rate with lexical features. 

\begin{table}[htb]
\begin{center}
\begin{tabular}{lr}
\toprule
   candidate marker &  accuracy ($\%$) \\
\midrule
  evidently, &       0.0 \\
 frequently, &       0.0 \\
   meantime, &       0.0 \\
 truthfully, &       0.0 \\
 supposedly, &       0.1 \\
\bottomrule
\end{tabular}
\end{center}
\caption{Candidate discourse markers that are the most difficult to predict from shallow features}
\label{table:lesspredictable}
\end{table}

\begin{table}[htb]
\begin{center}
\begin{tabular}{lr}
\toprule
    candidate marker &  accuracy  ($\%$)\\
\midrule
 defensively, &      65.5 \\
    afterward &      71.1 \\
  preferably, &      71.9 \\
        this, &      72.7 \\
        very, &      90.7 \\
\bottomrule
\end{tabular}
\end{center}
\caption{Candidate discourse markers that are the easiest to predict from shallow features. This shows candidates that are unlikely to be interesting discourse cues.}
\label{table:mostpredictable}
\end{table}

Interestingly, the two most predictable candidates are not discourse markers. Upon inspection of harvested pairs, we noticed that even legitimate discourse markers can be guessed with relatively simple heuristics in numerous examples.
For example, $c=\textit{thirdly}$ is very likely to occur if $s_1$ contains $\textit{secondly}$. 
We use this information to optionally filter out such simple instances, as described in the next section.

\subsection{Dataset variations}

In the following, we call our method \textit{Discovery}.
We create several variations of the sentence pairs dataset. In \textit{DiscoveryHard}, we remove examples where the candidate marker was among the top 5 predictions in our Fasttext shallow model and keep only the $174$ candidate markers with a frequency of at least $10k$. 
Instances are then sampled randomly so that each marker appears exactly $10k$ times in the dataset.

Subsequently, the resulting set of discourse markers is also used in the other variations of our dataset.  \textit{DiscoveryBase} designates the dataset for which examples predicted with the Fasttext model were not removed.  In order to measure the extent to which the model makes use of the relation between $s_1$ and $s_2'$, we also create a \textit{DiscoveryShuffled} dataset, which is the \textit{DiscoveryBase} dataset subjected to the \textit{random shuffle} operation described previously. To isolate the contribution of our discovery method, the dataset {\it DiscoveryAdv} discards all discourse markers from PDTB that were not found by our method. Also, in order to measure the impact of label diversity, {\it Discovery10} uses $174k$ examples for each of the $10$ most frequent markers,\footnote{They are: {\it however, hence, moreover, additionally, nevertheless, furthermore, alternatively, again, next, therefore}} thus totalling as many instances as {\it DiscoveryBase}.
Finally, \textit{DiscoveryBig} contains almost twice as many instances as {\it DiscoveryBase}, i.e. $20k$ instances for each discourse marker (although, for a limited number of markers, the number of instances is slightly lower due to data sparseness).

\begin{table*}[htb]
\begin{small}
\centering
\begin{tabular}{lrllllllllll}
\toprule
{} &      $N$ &               MR &                  CR &             SUBJ &             MPQA &             SST2 &             TREC &     SICK-R &   SICK-E &                MRPC &               AVG \\
\midrule
InferSent         &    1.0 &             81.1 &                86.3 &             92.4 &             90.2 &             84.6 &             88.2 &                88.4 &             86.1 &                76.2 &                85.9 \\
MTL               &  124 &             82.5 &     $\textbf{87.7}$ &               94 &             90.9 &             83.2 &               93 &     $\textbf{88.8}$ &  $\textbf{87.8}$ &     $\textbf{78.6}$ &     $\textbf{87.4}$ \\
\midrule
SkipThought       &   74&             76.5 &                80.1 &             93.6 &             87.1 &               82 &             92.2 &                85.8 &             82.3 &                  73 &                83.6 \\
QuickThought      &  174 &             81.3 &                84.5 &  $\textbf{94.6}$ &             89.5 &              - &             92.4 &  $\underline{87.1}$ &              - &                75.9 &               - \\
DisSent           &    4.7 &             80.1 &                84.9 &             93.6 &             90.1 &             84.1 &  $\textbf{93.6}$ &                84.9 &             83.7 &                  75 &                85.6 \\
DiscoveryBase         &    1.7 &             82.5 &                86.3 &             94.2 &             90.5 &             85.2 &             91.8 &                85.7 &               84 &                75.8 &                86.2 \\
DiscoveryHard     &    1.7 &             81.6 &                86.5 &             93.9 &             90.5 &             84.8 &               90 &                85.4 &             83.2 &                76.5 &                85.8 \\
Discovery10       &    1.7 &             81.2 &                85.1 &             93.7 &             90.2 &               83 &               90 &                85.9 &             83.8 &                75.8 &                85.4 \\
DiscoveryAdv      &    1.4 &             81.4 &                85.8 &             93.8 &             90.5 &             83.4 &               92 &                  86 &             84.3 &                75.7 &                85.9 \\
DiscoveryShuffled &    1.7 &             81.4 &                86.1 &             94.1 &             90.9 &  $\textbf{85.3}$ &             90.4 &                85.6 &             83.6 &                75.4 &                85.9 \\
DiscoveryBig      &    3.4 &  $\textbf{82.6}$ &  $\underline{87.4}$ &             94.5 &  $\textbf{91.0}$ &             85.2 &             93.4 &                86.4 &             \underline{84.8} &  $\underline{76.6}$ &  $\underline{86.9}$ \\
\bottomrule
\end{tabular}

 \caption{SentEval evaluation results with our models trained on various datasets.
 The first two models are supervised, the other ones unsupervised. All scores are accuracy percentages, except SICK-R, which is Pearson correlation percentage.
 InferSent is from  \citet{Conneau2017},
 MTL is the multi-task learning based model from \citet{subramanian2018learning}.
 Evaluation tasks are described in table \ref{table:evaltasks}, and $N$ denotes the number of examples for each dataset (in millions).
 Dissent is from \citet{Nie2017}, QuickThought is from \citet{Logeswaran2018} with fixed embeddings configuration. 
 The best result per task appears in bold, the best result for unsupervised setups is underlined.
 \label{table:ressenteval}
 }
\end{small}
\end{table*}



\section{Evaluation of sentence representation learning}
\subsection{Setup}
\begin{figure*}[htb]
  \centering
\includegraphics[angle=0, width=1\textwidth, trim=5 5 5 5,clip ]{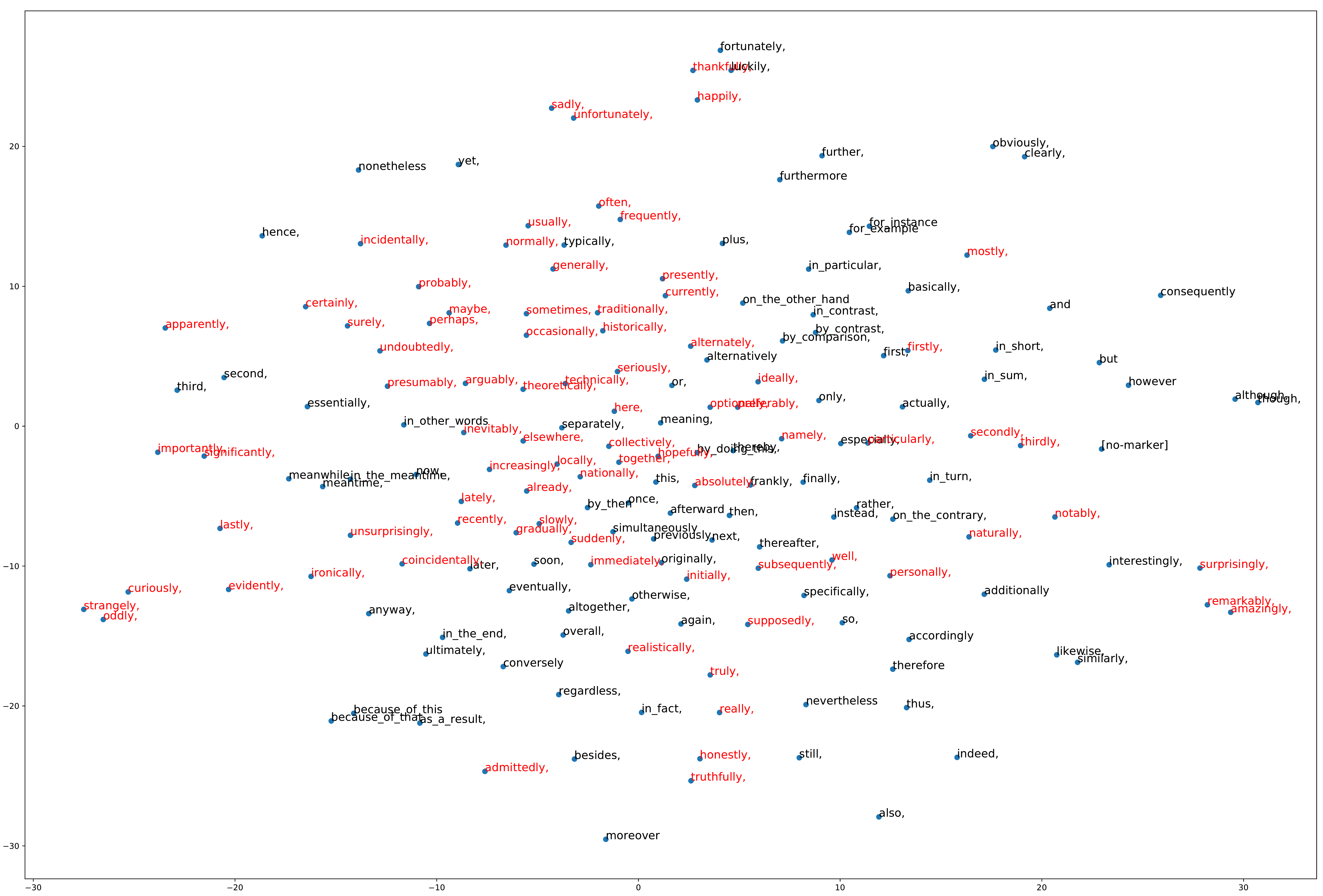}
\caption{TSNE visualization of the softmax weights from our \textit{DiscoveryBig} model for each discourse marker, after unit norm normalization. Markers discovered by our method (e.g. absent from PDTB annotations) are colored in red.}
\label{fig:tsne} 
\end{figure*}

Our goal is to evaluate the effect of using our various training datasets on sentence encoding, given encoders of equivalent capacity and similar setups.
Thus, we follow the exact setup of Infersent \citep{Conneau2017}, also used in the Dissent \citep{Malmi2018} model: we learn to encode sentences into $h$ with a bi-directional LSTM  sentence encoder using element-wise max pooling over time. 
The dimension size of $h$ is $4096$.  Word embeddings are fixed GloVe embeddings with 300 dimensions, trained on Common Crawl 840B.\footnote{https://nlp.stanford.edu/projects/glove/} 
A sentence pair ($s_1$, $s_2$) is represented with $ [h_1, h_2, h_1 \odot h_2, |h_2-h_1|]$,\footnote{$h_1 \odot h_2 = (h_{11}.h_{21},..,h_{1i}.h_{2i},...)$} which is fed to a softmax in order to predict a marker $c$.
Our datasets are split in $90\%$ train, $5\%$ validation, and $5\%$ test. 
Optimization is done with SGD (learning rate is initialized at $0.1$, decayed by $1\%$ at each epoch and by $80\%$ if validation accuracy decreases; learning stops when learning rate is below $10^{-5}$ and the best model on training task validation loss is used for evaluation; gradient is clipped when its norm exceeds $5$). Once the sentence encoder has been trained on a base task, the resulting sentence embeddings are tested with the SentEval library \citep{Conneau2017}.

We evaluate the different variations of our dataset we described above in order to analyze their effect, and compare them to a number of existing models. Table \ref{table:evaltasks} displays the tasks used for evaluation.
For further analysis, table \ref{tab:probe} displays the result of {\it Linguistic Probing} using the method by \citet{ConneauProbe}.  Although these tasks are primarily designed for understanding the content of embeddings, they also focus on aspects that are desirable to perform well in general semantic tasks (e.g. prediction of tense, or number of object).

\subsection{Results}
Table \ref{table:ressenteval} gives an overview of transfer learning evaluation, also comparing to other supervised and unsupervised approaches.
Note that we outperform DisSent on all tasks except TREC\footnote{This dataset is composed of questions only, which are underrepresented in our training data.} with less than half the amount of training examples. In addition, our approach is arguably simpler and faster.

MTL \cite{subramanian2018learning} only achieves stronger results than our method on the MRPC and SICK tasks. The MTL model uses $124M$ training examples with an elaborate multi-task setup, training on $45M$ sentences with manual translation, $1M$ pairs from SNLI/MNLI, $4M$ parse trees of sentences, and $74M$ consecutive sentence pairs. The model also fine-tunes word embeddings in order to achieve a higher capacity. It is therefore remarkable that our model outperforms it on many tasks. Besides, MTL is not a direct competitor to our approach since its main contribution is its multi-task setup, and it could benefit from using our training examples. 

Our best model rivals (and indeed often outperforms) QuickThought on all tasks, except relatedness (SICK-R). QuickThought's training task is to predict whether two sentences are contiguous, which might incentivize the model to perform well on a relatedness task. We also outperform InferSent on many tasks except entailment and relatedness. Entailment prediction is the explicit training signal for Infersent.

To help the analysis of our different model variations, table \ref{table:disctasks} displays the test scores on each dataset for the original training task. It also shows the related PDTB implicit relation prediction scores. 
The PDTB is annotated with a hierarchy of relations, with 5 classes at level 1 (including the EntRel relation), and 16 at level 2 (with one relation absent from the test). 
It is interesting to see that this form of simple semi-supervised learning for implicit relation prediction performs quite well, especially for fine-grained relations, as the best model slightly beats the best current dedicated model, listed at 40.9\% in \citet{DBLP:conf/eacl/XueDR17}. 

{\it DiscoveryHard} scores lower on its training task than {\it DiscoveryBase}, and it also performs worse on transfer learning tasks. This makes sense, since lexical features are important to solve the evaluation tasks. Our initial hypothesis was that more difficult instances might force the model to use higher-level reasoning, but this does not seem to be the case. More surprisingly, preventing the encoders to use the relationship between sentences, as in {\it DiscoveryShuffled}, does not substantially hurt the transfer performance, which remains on average higher than \citet{Nie2017}. Additionally,  our models score well on linguistic probing tasks. They outperform Infersent on all tasks, which seems to contradict the claim that SNLI data allows for learning of {\it universal} sentence representations \citep{Conneau2017}. And a final interesting outcome is that the diversity of markers (e.g. using {\it DiscoveryBase} instead of {\it Discovery10})  seems to be important for good performance on those tasks, since {\it Discovery10} has the worst overall performance on average.

 \begin{table}[htb]
\small
\begin{center}
\begin{tabular}{lllll}
\toprule
   name &     N &                         task &           C              \\
\midrule
     MR &   11k &           sentiment (movie reviews) &           2 &                        \\
     CR &    4k &  sentiment (product reviews) &           2 &                       \\
   SUBJ &   10k &     subjectivity/objectivity &           2 &                         \\
   MPQA &   11k &             opinion polarity &           2 & \\
   TREC &    6k &                question-type &           6 &                     \\
   SST &   70k &                sentiment (movie reviews) &           2 &                        \\
 SICK-E&   10k &             entailment  &           3 &         \\
  SICK-R&   10k &                          relatedness &           3 &             \\
   MRPC &  4k &         paraphrase detection &           2 &   \\
  $\text{PDTB}_5$ &  17k &                 implicit discourse relation (coarse) &  5 &         \\
  $\text{PDTB}_{16}$ &  17k &                 implicit discourse relation (fine) &  15 &         \\

\bottomrule
\end{tabular}
\end{center}
 \caption{Transfer evaluation tasks. N is the number of training examples and C is number of classes for each task.}
\label{table:evaltasks}
 \end{table}

 \begin{table}[htb]
\begin{small}
\begin{center}
\begin{tabular}{lp{1cm}p{1cm}r}
\toprule
{} &  $\text{PDTB}_5$ coarse&  $\text{PDTB}_{16}$ fine&   $\mathcal T$ \\
\midrule
InferSent          &  46.7 &   34.2 &  - \\
DisSent & 48.9& 36,9 & - \\
DiscoveryBase         &  \textbf{52.5} &   40.0 &  20.6 \\
DiscoveryHard     &  50.7 &   39.8 &   9.3 \\
Discovery10       &  48.3 &   37.7 &  51.9 \\
DiscoveryAdv      &  49.7 &   37.6 &  26.1 \\
DiscoveryShuffled &  51.0 &   39.5 &  11.5 \\
DiscoveryBig      &  51.3 &   \textbf{41.3} &  22.2 \\
\bottomrule
\end{tabular}
\end{center}
\caption{Test results (accuracy) on implicit discursive relation prediction task (PDTB relations level 1 and 2, i.e coarse-grained and fine-grained) and training tasks $\mathcal T$. Note that scores for $\mathcal T$ are not comparable since the test set changes for each version of the dataset. }
\label{table:disctasks}
\end{small}
\end{table}

\subsection{Visualisation}
The softmax weights learned during the training phase can be interpreted as embeddings for the markers themselves, and used to visualize their relationships.
Figure \ref{fig:tsne} shows a TSNE \citep{vanDerMaaten2008} plot of the markers' representations. Proximity in the feature space seems to reflect semantic similarity (e.g. \textit{usually}/\textit{normally}). In addition, the markers we discovered, colored in red, blend 
with the PDTB markers (depicted in black).
It would be interesting to cluster markers in order to empirically define discourse relations, but we leave this for future work. 
\begin{table*}[htb]
\centering
\small
\begin{tabular}{lllllllllll}
\toprule
{} &      BShift & CoordInv&            Depth &        ObjNum &       SubjNum &        OddM &            Tense &  TC&      WC&              AVG \\
\midrule
InferSent         &             56.5 &                  65.9 &             37.5 &             79.9 &             84.3 &             53.2 &               87 &             78.1 &             95.2 &             70.8 \\
SkipThought       &  $\textbf{69.5}$ &                    69 &             39.6 &             83.2 &             86.2 &             54.5 &  $\textbf{90.3}$ &             82.1 &             79.6 &             72.7 \\
QuickThought      &             56.8 &                    70 &             40.2 &             79.7 &               83 &             55.3 &             86.2 &             80.7 &             90.3 &             71.4 \\
DiscoveryBase         &             63.1 &                  70.6 &             45.2 &             83.8 &             87.2 &             57.3 &             89.1 &             83.2 &             94.7 &             74.9 \\
DiscoveryHard     &             62.7 &                  70.4 &             44.5 &             83.4 &  $\textbf{88.1}$ &             57.3 &             89.5 &             82.8 &             94.1 &             74.8 \\
Discovery10       &             61.3 &                  69.7 &             42.9 &             81.8 &             86.7 &             55.8 &             87.8 &             81.4 &  $\textbf{96.1}$ &             73.7 \\
DiscoveryAdv      &             61.5 &                    70 &             43.9 &             82.6 &             86.2 &             56.2 &             89.1 &             82.8 &  $\textbf{96.1}$ &             74.3 \\
DiscoveryShuffled &             62.6 &       $\textbf{71.4}$ &             45.3 &  $\textbf{84.3}$ &               88 &  $\textbf{58.3}$ &             89.3 &             82.8 &             93.4 &               75 \\
DiscoveryBig      &             63.3 &       $\textbf{71.4}$ &  $\textbf{46.0}$ &             84.1 &             87.8 &             57.1 &             89.4 &  $\textbf{84.2}$ &               96 &  $\textbf{75.5}$ \\
\bottomrule
\end{tabular}

\caption{
Accuracy of various models on linguistic probing tasks using logistic regression on SentEval.
BShift is detection of token inversion.  CoordInv is detection of clause inversion. ObjNum/SubjNum is prediction of the number of object resp. subject. Tense is prediction of the main verb tense. Depth is prediction of parse tree depth. TC is detection of common sequences of constituents. WC is prediction of words contained in the sentence. OddM is detection of random replacement of verbs/nouns by other verbs/nouns. AVG is the average score of those tasks for each model.
For more details see \citet{ConneauProbe}. SkipThought and Infersent results come from \citet{Perone2018EvaluationOS}, QuickThought results come from \citet{Brahma2018UnsupervisedLO}.}
\label{tab:probe}
\end{table*}
\section{Related work}
Though discourse marker prediction in itself is an interesting and useful task \citep{Malmi2017}, discourse markers have often been used as a training cue in order to improve implicit relation prediction \citep{Marcu2001,Sporleder2005ExploitingLC,Zhou2010,braud-denis:2016:EMNLP2016}.
This approach has been extended to general representation learning by \citet{Jernite2017}---although with empirically unconvincing results, which might be attributed to an inappropriate training/evaluation set-up, or the use of a limited number of broad categories instead of actual discourse markers. \citet{Nie2017} used the more standard InferSent framework and obtained better results, although they were still outperformed by QuickThought \citep{Logeswaran2018}, which uses a much simpler training task. Both of these rely on pre-established lists of discourse markers provided by the PDTB, and both perform a manual annotation for each marker---\citet{Nie2017} uses dependency patterns, while \citet{Jernite2017} uses broad discourse categories. Our work is the first to automatically discover discourse markers from text.

More generally, various automatically extracted training signals have been used for unsupervised learning tasks. Hashtags \citep{Felbo2017UsingMO} have been sucessfully exploited in order to learn sentiment analysis from unlabelled tweets, but their availability is mainly limited to the microblogging domain.
Language modeling provides a general training signal for representation learning, even though there is no obvious way to derive sentence representations from language models. BERT \citep{devlin2018bert} currently holds the best results in transfer learning based on language modeling, but it relies on sentence pair classification in order to compute sentence embeddings, and it makes use of a simple sentence contiguity detection task (like QuickThought); this task does not seem challenging enough since BERT reportedly achieves 98\% detection accuracy. \citet{PhangSTILTS} showed that the use of SNLI datasets yields significant gains for the sentence embeddings from \citet{Radford2018ImprovingLU}, which are based on language modeling.

For the analysis of our models, we draw inspiration from critical work on Natural Language Inference datasets \citep{Dasgupta2018, Levy2018}.  
\citet{Gururangan2018,Poliak2018} show that baseline models that disregard the hypothesis yield good results on SNLI, which suggests that the model does not perform the high level reasoning we would expect in order to predict the correct label. They attribute this effect to bias in human annotations. In this work, we show that this issue is not inherent to human labeled data, and propose the \textit{shuffle} perturbation in order to measure to what extent the relationship between sentences is used.

\section{Conclusion}
In this paper, we introduce a novel and efficient method to automatically discover discourse markers from text, and we use the resulting set of candidate markers for the construction of an extensive dataset for semi-supervised sentence representation learning. A number of dataset variations are evaluated on a wide range of transfer learning tasks (as well as implicit discourse recognition) and a comparison with existing models indicates that our approach yields state of the art results on the bulk of these tasks. Additionally, our analysis shows that removing `simple' examples is detrimental to transfer results, while preventing the model to exploit the relationship between sentences has a negligible effect. This leads us to believe that, even though our approach reaches state of the art results, there is still room for improvement: models that adequately exploit the relationship between sentences would be better at leveraging the supervision of our dataset, and could yield even better sentence representations.
In future work, we also aim to increase the coverage of our method. For instance, we can make use of more lenient patterns that capture an even wider range of discourse markers, such as multi-word markerse.



\bibliographystyle{acl_natbib}
\bibliography{naaclhlt2019}

\begin{thebibliography}{39}
\expandafter\ifx\csname natexlab\endcsname\relax\def\natexlab#1{#1}\fi

\bibitem[{Barzilay and Lapata(2008)}]{Barzilay2008}
Regina Barzilay and Mirella Lapata. 2008.
\newblock \href {https://doi.org/10.1162/coli.2008.34.1.1} {{Modeling Local
  Coherence: An Entity-Based Approach}}.
\newblock \emph{Computational Linguistics}, 34(1):1--34.

\bibitem[{Bird et~al.(2009)Bird, Klein, and Loper}]{Bird2009}
Steven Bird, Ewan Klein, and Edward Loper. 2009.
\newblock \href {https://doi.org/10.1097/00004770-200204000-00018}
  {\emph{{Natural Language Processing with Python}}}, volume~43.

\bibitem[{Brahma(2018)}]{Brahma2018UnsupervisedLO}
Siddhartha Brahma. 2018.
\newblock Unsupervised learning of sentence representations using sequence
  consistency.
\newblock \emph{CoRR}, abs/1808.04217.

\bibitem[{Braud and Denis(2016)}]{braud-denis:2016:EMNLP2016}
Chlo\'{e} Braud and Pascal Denis. 2016.
\newblock \href {https://aclweb.org/anthology/D16-1020} {Learning
  connective-based word representations for implicit discourse relation
  identification}.
\newblock In \emph{Proceedings of the 2016 Conference on Empirical Methods in
  Natural Language Processing}, pages 203--213, Austin, Texas. Association for
  Computational Linguistics.

\bibitem[{Carlson et~al.(2001)Carlson, Marcu, and
  Okurowski}]{DBLP:conf/sigdial/CarlsonMO01}
Lynn Carlson, Daniel Marcu, and Mary~Ellen Okurowski. 2001.
\newblock \href {https://doi.org/10.3115/1118078.1118083} {Building a
  discourse-tagged corpus in the framework of rhetorical structure theory}.
\newblock In \emph{Proceedings of the Second SIGdial Workshop on Discourse and
  Dialogue - Volume 16}, SIGDIAL '01, pages 1--10, Stroudsburg, PA, USA.
  Association for Computational Linguistics.

\bibitem[{Conneau et~al.(2017)Conneau, Kiela, Schwenk, Barrault, and
  Bordes}]{Conneau2017}
Alexis Conneau, Douwe Kiela, Holger Schwenk, Lo{\"{\i}}c Barrault, and Antoine
  Bordes. 2017.
\newblock Supervised learning of universal sentence representations from
  natural language inference data.
\newblock In \emph{{EMNLP}}, pages 670--680. Association for Computational
  Linguistics.

\bibitem[{Conneau et~al.(2018)Conneau, Kruszewski, Lample, Barrault, and
  Baroni}]{ConneauProbe}
Alexis Conneau, Germ{\'a}n Kruszewski, Guillaume Lample, Lo{\"i}c Barrault, and
  Marco Baroni. 2018.
\newblock \href {http://aclweb.org/anthology/P18-1198} {What you can cram into
  a single vector: Probing sentence embeddings for linguistic properties}.
\newblock In \emph{Proceedings of the 56th Annual Meeting of the Association
  for Computational Linguistics (Volume 1: Long Papers)}, pages 2126--2136.
  Association for Computational Linguistics.

\bibitem[{Dasgupta et~al.(2018)Dasgupta, Guo, Stuhlm{\"{u}}ller, Gershman, and
  Goodman}]{Dasgupta2018}
Ishita Dasgupta, Demi Guo, Andreas Stuhlm{\"{u}}ller, Samuel~J. Gershman, and
  Noah~D. Goodman. 2018.
\newblock \href {http://arxiv.org/abs/1802.04302} {{Evaluating Compositionality
  in Sentence Embeddings}}.
\newblock (2011).

\bibitem[{Devlin et~al.(2018)Devlin, Chang, Lee, and
  Toutanova}]{devlin2018bert}
Jacob Devlin, Ming-Wei Chang, Kenton Lee, and Kristina Toutanova. 2018.
\newblock Bert: Pre-training of deep bidirectional transformers for language
  understanding.
\newblock \emph{arXiv preprint arXiv:1810.04805}.

\bibitem[{Felbo et~al.(2017)Felbo, Mislove, Sogaard, Rahwan, and
  Lehmann}]{Felbo2017UsingMO}
Bjarke Felbo, Alan Mislove, Anders Sogaard, Iyad Rahwan, and Sune Lehmann.
  2017.
\newblock Using millions of emoji occurrences to learn any-domain
  representations for detecting sentiment, emotion and sarcasm.
\newblock In \emph{EMNLP}.

\bibitem[{Glockner et~al.(2018)Glockner, Shwartz, and Goldberg}]{Glockner2018}
Max Glockner, Vered Shwartz, and Yoav Goldberg. 2018.
\newblock \href {http://arxiv.org/abs/1805.02266} {{Breaking NLI Systems with
  Sentences that Require Simple Lexical Inferences}}.
\newblock \emph{Proceedings of the 56th Annual Meeting of the Association for
  Computational Linguistics (Short Papers)}, (3):1--6.

\bibitem[{Grave et~al.(2018)Grave, Bojanowski, Gupta, Joulin, and
  Mikolov}]{GRAVE18.627}
Edouard Grave, Piotr Bojanowski, Prakhar Gupta, Armand Joulin, and Tomas
  Mikolov. 2018.
\newblock {Learning Word Vectors for 157 Languages}.
\newblock In \emph{Proceedings of the Eleventh International Conference on
  Language Resources and Evaluation (LREC 2018)}, Miyazaki, Japan. European
  Language Resources Association (ELRA).

\bibitem[{Gururangan et~al.(2018)Gururangan, Swayamdipta, Levy, Schwartz,
  Bowman, and Smith}]{Gururangan2018}
Suchin Gururangan, Swabha Swayamdipta, Omer Levy, Roy Schwartz, Samuel~R.
  Bowman, and Noah~A. Smith. 2018.
\newblock \href {http://arxiv.org/abs/1803.02324} {{Annotation Artifacts in
  Natural Language Inference Data}}.

\bibitem[{Jernite et~al.(2017)Jernite, Bowman, and Sontag}]{Jernite2017}
Yacine Jernite, Samuel~R. Bowman, and David Sontag. 2017.
\newblock \href {http://arxiv.org/abs/1705.00557} {{Discourse-Based Objectives
  for Fast Unsupervised Sentence Representation Learning}}.

\bibitem[{Joulin et~al.(2016)Joulin, Grave, Bojanowski, and
  Mikolov}]{joulin2016tricks}
Armand Joulin, Edouard Grave, Piotr Bojanowski, and Tomas Mikolov. 2016.
\newblock \href {http://arxiv.org/abs/1607.01759} {Bag of tricks for efficient
  text classification}.
\newblock Cite arxiv:1607.01759.

\bibitem[{Kiros et~al.(2015)Kiros, Zhu, Salakhutdinov, Zemel, Urtasun,
  Torralba, and Fidler}]{Kiros2015}
Ryan Kiros, Yukun Zhu, Ruslan~R Salakhutdinov, Richard Zemel, Raquel Urtasun,
  Antonio Torralba, and Sanja Fidler. 2015.
\newblock {Skip-thought vectors}.
\newblock In \emph{Advances in neural information processing systems}, pages
  3294--3302.

\bibitem[{Levy et~al.(2018)Levy, Bowman, and Smith}]{Levy2018}
Omer Levy, Samuel~R Bowman, and Noah~A Smith. 2018.
\newblock {Annotation Artifacts in Natural Language Inference Data}.
\newblock \emph{Proceedings of NAACL-HLT 2018}, pages 107--112.

\bibitem[{Logeswaran and Lee(2018)}]{Logeswaran2018}
Lajanugen Logeswaran and Honglak Lee. 2018.
\newblock \href {http://arxiv.org/abs/1803.02893} {{An efficient framework for
  learning sentence representations}}.
\newblock pages 1--16.

\bibitem[{Logeswaran et~al.(2016)Logeswaran, Lee, and Radev}]{Logeswaran2016}
Lajanugen Logeswaran, Honglak Lee, and Dragomir Radev. 2016.
\newblock \href {http://arxiv.org/abs/1611.02654} {{Sentence Ordering using
  Recurrent Neural Networks}}.
\newblock pages 1--15.

\bibitem[{van~der Maaten and Hinton(2008)}]{vanDerMaaten2008}
Laurens van~der Maaten and Geoffrey Hinton. 2008.
\newblock \href {http://www.jmlr.org/papers/v9/vandermaaten08a.html}
  {Visualizing data using {t-SNE}}.
\newblock \emph{Journal of Machine Learning Research}, 9:2579--2605.

\bibitem[{Malmi et~al.(2017)Malmi, Pighin, Krause, and Kozhevnikov}]{Malmi2017}
Eric Malmi, Daniele Pighin, Sebastian Krause, and Mikhail Kozhevnikov. 2017.
\newblock \href {http://arxiv.org/abs/1702.00992} {{Automatic Prediction of
  Discourse Connectives}}.

\bibitem[{Malmi et~al.(2018)Malmi, Pighin, Krause, and Kozhevnikov}]{Malmi2018}
Eric Malmi, Daniele Pighin, Sebastian Krause, and Mikhail Kozhevnikov. 2018.
\newblock {Automatic Prediction of Discourse Connectives}.
\newblock In \emph{Proceedings of the Eleventh International Conference on
  Language Resources and Evaluation (LREC 2018)}, Miyazaki, Japan. European
  Language Resources Association (ELRA).

\bibitem[{Marcu and Echihabi(2001)}]{Marcu2001}
Daniel Marcu and Abdessamad Echihabi. 2001.
\newblock \href {https://doi.org/10.3115/1073083.1073145} {{An unsupervised
  approach to recognizing discourse relations}}.
\newblock \emph{Proceedings of the 40th Annual Meeting on Association for
  Computational Linguistics - ACL '02}, (July):368.

\bibitem[{Marcu and Echihabi(2002)}]{marcu2002unsupervised}
Daniel Marcu and Abdessamad Echihabi. 2002.
\newblock An unsupervised approach to recognizing discourse relations.
\newblock In \emph{Proceedings of the 40th Annual Meeting on Association for
  Computational Linguistics}, pages 368--375.

\bibitem[{Nie et~al.(2017)Nie, Bennett, and Goodman}]{Nie2017}
Allen Nie, Erin~D. Bennett, and Noah~D. Goodman. 2017.
\newblock \href {http://arxiv.org/abs/1710.04334} {{DisSent: Sentence
  Representation Learning from Explicit Discourse Relations}}.

\bibitem[{Panchenko et~al.(2017)Panchenko, Ruppert, Faralli, Ponzetto, and
  Biemann}]{Panchenko2017}
Alexander Panchenko, Eugen Ruppert, Stefano Faralli, Simone~P Ponzetto, and
  Chris Biemann. 2017.
\newblock {Building a Web-Scale Dependency-Parsed Corpus from Common Crawl}.
\newblock pages 1816--1823.

\bibitem[{Perone et~al.(2018)Perone, Silveira, and
  Paula}]{Perone2018EvaluationOS}
Christian~S. Perone, Roberto Silveira, and Thomas~S. Paula. 2018.
\newblock Evaluation of sentence embeddings in downstream and linguistic
  probing tasks.
\newblock \emph{CoRR}, abs/1806.06259.

\bibitem[{Phang et~al.(2018)Phang, F{\'{e}}vry, and Bowman}]{PhangSTILTS}
Jason Phang, Thibault F{\'{e}}vry, and Samuel~R. Bowman. 2018.
\newblock \href {http://arxiv.org/abs/1811.01088} {Sentence encoders on stilts:
  Supplementary training on intermediate labeled-data tasks}.
\newblock \emph{CoRR}, abs/1811.01088.

\bibitem[{Pitler and Nenkova(2009)}]{DBLP:conf/acl/PitlerN09}
Emily Pitler and Ani Nenkova. 2009.
\newblock \href {http://aclweb.org/anthology/P09-2004} {Using syntax to
  disambiguate explicit discourse connectives in text}.
\newblock In \emph{Proceedings of the ACL-IJCNLP 2009 Conference Short Papers},
  pages 13--16. Association for Computational Linguistics.

\bibitem[{Pitler et~al.(2008)Pitler, Raghupathy, Mehta, Nenkova, Lee, and
  Joshi}]{pitler2008}
Emily Pitler, Mridhula Raghupathy, Hena Mehta, Ani Nenkova, Alan Lee, and
  Aravind Joshi. 2008.
\newblock \href {http://aclweb.org/anthology/C08-2022} {Easily identifiable
  discourse relations}.
\newblock In \emph{Coling 2008: Companion volume: Posters}, pages 87--90.
  Coling 2008 Organizing Committee.

\bibitem[{Poliak et~al.(2018)Poliak, Naradowsky, Haldar, Rudinger, and
  Durme}]{Poliak2018}
Adam Poliak, Jason Naradowsky, Aparajita Haldar, Rachel Rudinger, and
  Benjamin~Van Durme. 2018.
\newblock {Hypothesis Only Baselines in Natural Language Inference}.
\newblock \emph{Proceedings of the 7th Joint Conference on Lexical and
  Computational Semantics}, (1):180--191.

\bibitem[{Prasad et~al.(2008)Prasad, Dinesh, Lee, Miltsakaki, Robaldo, Joshi,
  and Webber}]{pdtb2.0}
Rashmi Prasad, Nikhil Dinesh, Alan Lee, Eleni Miltsakaki, Livio Robaldo,
  Aravind Joshi, and Bonnie Webber. 2008.
\newblock The penn discourse treebank 2.0.
\newblock In \emph{Proceedings of the Sixth International Conference on
  Language Resources and Evaluation (LREC'08)}, Marrakech, Morocco. European
  Language Resources Association (ELRA).
\newblock Http://www.lrec-conf.org/proceedings/lrec2008/.

\bibitem[{Radford(2018)}]{Radford2018ImprovingLU}
Alec Radford. 2018.
\newblock Improving language understanding by generative pre-training.

\bibitem[{Rutherford et~al.(2017)Rutherford, Demberg, and
  Xue}]{DBLP:conf/eacl/XueDR17}
Attapol Rutherford, Vera Demberg, and Nianwen Xue. 2017.
\newblock \href {http://aclweb.org/anthology/E17-1027} {A systematic study of
  neural discourse models for implicit discourse relation}.
\newblock In \emph{Proceedings of the 15th Conference of the European Chapter
  of the Association for Computational Linguistics: Volume 1, Long Papers},
  pages 281--291. Association for Computational Linguistics.

\bibitem[{Rutherford and Xue(2015)}]{RutherfordX15}
Attapol Rutherford and Nianwen Xue. 2015.
\newblock Improving the inference of implicit discourse relations via
  classifying explicit discourse connectives.
\newblock In \emph{{HLT-NAACL}}, pages 799--808. The Association for
  Computational Linguistics.

\bibitem[{Sporleder and Lascarides(2005)}]{Sporleder2005ExploitingLC}
Caroline Sporleder and Alex Lascarides. 2005.
\newblock Exploiting linguistic cues to classify rhetorical relations.
\newblock In \emph{Proceedings of Recent Advances in Natural Langauge
  Processing (RANLP)}, Bulgaria.

\bibitem[{Sporleder and Lascarides(2008)}]{DBLP:journals/nle/SporlederL08}
Caroline Sporleder and Alex Lascarides. 2008.
\newblock \href {https://doi.org/10.1017/S1351324906004451} {Using
  automatically labelled examples to classify rhetorical relations: an
  assessment}.
\newblock \emph{Natural Language Engineering}, 14(3):369--416.

\bibitem[{Subramanian et~al.(2018)Subramanian, Trischler, Bengio, and
  Pal}]{subramanian2018learning}
Sandeep Subramanian, Adam Trischler, Yoshua Bengio, and Christopher~J Pal.
  2018.
\newblock Learning general purpose distributed sentence representations via
  large scale multi-task learning.
\newblock \emph{International Conference on Learning Representations}.

\bibitem[{Zhou et~al.(2010)Zhou, Xu, Niu, Lan, Su, and Tan}]{Zhou2010}
Zhi-Min Zhou, Yu~Xu, Zheng-Yu Niu, Man Lan, Jian Su, and Chew~Lim Tan. 2010.
\newblock \href {http://aclweb.org/anthology/C10-2172} {Predicting discourse
  connectives for implicit discourse relation recognition}.
\newblock In \emph{Coling 2010: Posters}, pages 1507--1514. Coling 2010
  Organizing Committee.

\end{thebibliography}

\end{document}